\documentclass{article}
\usepackage{ijcai19}
\usepackage{times}
\usepackage{soul}
\usepackage[utf8]{inputenc}
\usepackage[small]{caption}
\usepackage{graphicx}
\usepackage{amsmath,amssymb,amsthm}
\usepackage{booktabs}
\usepackage{algorithm}
\usepackage{algorithmic}
\usepackage[usenames, dvipsnames]{xcolor}
\usepackage{xspace}
\usepackage[switch]{lineno}
\usepackage{lipsum}
\usepackage{comment}
\usepackage{mathrsfs}
\usepackage{url}

\title{
LogicENN: A Neural Based  Knowledge Graphs Embedding Model\\ with Logical Rules}

\author{}

\newtheorem{theorem}{Theorem}

\usepackage{etoolbox}
\usepackage{bm}
\renewcommand{\vec}[2][]{
  \ensuremath
  \ifstrempty{#1}{
    \bm{\lowercase{#2}}
  }{%
    [\bm{\lowercase{#2}}]_{#1}
  }%
}

\newcommand{\mat}[2][]{
  \ensuremath
  \ifstrempty{#1}{
    \bm{\uppercase{#2}}
  }{%
    [\bm{\uppercase{#2}}]_{#1}
  }%
}

\newcommand{\ten}[2][]{
  \ensuremath
  \ifstrempty{#1}{
    \bm{\underline{\uppercase{#2}}}
  }{%
    [\bm{\underline{\uppercase{#2}}}]_{#1}
  }%
}

\newcommand{\name}{LogicENN\xspace}

\newcommand{\var}[1]{\mathtt{#1}}

\author{
Mojtaba Nayyeri\footnote{This work is the arxiv version of the paper uploaded in openreview}$^1$
\and
Chengjin Xu$^1$\and
Jens Lehmann$^{1,2}$\And
Hamed Shariat Yazdi$^1$
\affiliations
$^1$University of Bonn, Bonn, Germany\\
$^2$ Fraunhofer IAIS, Bonn, Germany\\
\emails
\{nayyeri, xu, jens.lehmann, Shariat\}@cs.uni-bonn.de,
}

\begin{document}

\maketitle
  
\begin{abstract}
Knowledge graph embedding models have gained significant attention in AI research.
Recent works have shown that the inclusion of background knowledge, such as logical rules, can improve the performance of embeddings in downstream machine learning tasks.
However, so far, most existing models do not allow the inclusion of rules.  
We address the challenge of including rules and present a new neural based embedding model (\name).
We prove that \name can learn every ground truth of encoded rules in a knowledge graph.
To the best of our knowledge, this has not been proved so far for the neural based family of embedding models.
Moreover, we derive formulae for the inclusion of various rules, including (anti-)symmetric, inverse, irreflexive and transitive, implication, composition, equivalence and negation.
Our formulation allows to avoid grounding for implication and equivalence relations. 
Our experiments show that \name outperforms the state-of-the-art models in link prediction. 
\end{abstract}

\section{Introduction}
\label{sec:Intro}

Knowledge graphs (KGs) such as DBpedia, Freebase and Yago encode structured information in the form of a multi-relational directed graph in which nodes represent entities and edges represent relations between nodes. 
In its simplest form, a KG is a collection of triples $(h,r,t)$ where $h$ and $t$ are head and tail entities (nodes), respectively, and $r$ is the relation (edge) between. For instance, (Albert Einstein, co-author, Boris Podolsky) is a triple.
Link prediction, entity resolution and linked-based clustering are among the most common tasks in KG analysis
\cite{Nickel2015ReviewRelationalMLforKG}.

Knowledge Graph Embeddings (KGEs) have become one of the most promising approaches for KG analysis \cite{wang2017surveyEmbeddings}.
The assumption is that there are global features which explain the existence of triples in a KG and embedding models try to capture those features using (typically low dimensional) vectors known as embeddings. Therefore, a KGE model assigns  vectors ($\vec{h}, \vec{r}, \vec{t}$) to the symbolic entities and relations  ($h, r, t$). The vectors are initialized randomly and updated by solving an optimization problem. To measure the degree of plausibility of a triple ($h,r,t$), a scoring function is defined. The function takes the embedding vectors of the triple and returns a value showing plausibility of the triple.
KGEs have a wide range of downstream applications such as recommender systems, question answering, sentiment analysis etc.

Several KGE models have been proposed so far.
Earlier works such as TransE \cite{bordes2013translating}, RESCAL \cite{nickel2012factorizing} and E-MLP \cite{socher2013reasoning} 
focus just on existing triples as inputs for the link prediction task (predicting missing relations between entities).
Due to the intrinsic incompleteness of KGs, relying only on triples may not deliver the best performance.
Recent works such as ComplEx-NNE+AER and RUGE have invested the usage of background knowledge such as logical rules in order to enhance the performance  \cite{guo2018RUGE,SimpConstding2018improving}. 

To exploit rules, the inherent \textit{incapability} of some existing models to encode rules is an obstacle.
For instance, different variants of translational approaches such as TransE, FTransE, STransE, TransH and TransR have restrictions in encoding reflexive, symmetric and transitivity relations \cite{SimplE}.
Considering TransE as a concrete example, the main intuition is that one should derive the embedding vector of tail $\vec{t}$ when it is the sum of the embedding vectors of head and relation i.e.~$\vec{h}+\vec{r}=\vec{t}$.
Once $r$ is assumed to be symmetric, e.g.\ ``co-author'', we have $\vec{t}+\vec{r}=\vec{h}$ which results in  $\vec{r}=\vec{0}$. Therefore, TransE cannot capture symmetry and collapses all symmetrical relations into a null vector, resulting the same embedding vectors of all entities (i.e.\ $\vec{h} = \vec{t}$).  

Due to incompleteness of KGs, even if one adds groundings of rules to a KG, there is still no guarantee that a capable embedding model learns the associated rules.
That means, we need to properly inject rules into the learning process of a capable model.
This issue has also been highlighted in the recent works \cite{guo2018RUGE,SimpConstding2018improving}, but not been investigated deeply in the literature.
Therefore, the \textbf{capability of a model} to support rules as well as how rules are injected, i.e.\ \textbf{encoding techniques}, are the main challenges.
Existing KGE models have solely addressed one of the mentioned challenges. This indeed causes two issues: 

a) Solely focusing on encoding techniques and disregarding capability of a model, has the risk that the model is expected to learn a rule which is not capable of. For example, RUGE proposes a general optimization framework to iteratively inject first order logical rules in an embedding model. ComplEx is used as the base model for rule injection. However, ComplEx is not capable of encoding composition pattern \cite{RotatE}. 
The similar issue can be found in \cite{minervini2017adversarial}. \citeauthor{minervini2017adversarial} use  function-free Horn clause rules to regularize an embedding model by including inconsistency loss. The loss measures degree of violation from assumption on adversarially generated examples. 
Although their framework nicely encodes Horn rules, they use DistMult to inject rules which is not capable of encoding asymmetric rule. For example, the authors \cite{minervini2017adversarial} try to inject $(h, Hypernym, t) \rightarrow (t, Hyponym, h)$ rule in DistMult. During injection of $(h, Hypernym, t) \rightarrow (t, Hyponym, h)$, $(h, Hypernym, t) \rightarrow (h, Hyponym, t)$ is also injected wrongly. Therefore, the model considers many false triples as positive. 

b) Solely focusing on capability of a model and disregarding encoding techniques, has the risk that the model does not properly encode rule due to incompleteness of KGs. For example, RotatE \cite{RotatE}
is proven to be capable of encoding inverse, symmetric (asymmetric), and composition rules without providing any rule injection mechanism. The authors \cite{RotatE} show their model properly encodes the rules. However, the results are obtained by generation of a lot of negative samples (e.g.~1000) together with using a very big embedding dimension (e.g.~1000). Such a big setting requires a very powerful computational infrastructure, adversely limit their applicability. Apart from lack of providing encoding technique, RotatE is not fully expressive i.e., the model is incapable of encoding some rules e.g., reflexive.

In contrast to the previous works, this paper addresses and contributes to the both previously highlighted points, i.e.\ the capability and the encoding technique, to avoid the mentioned issues.
Regarding capability, our first contribution is that we propose a new neural embedding model (\name) which is capable enough to encode rules, i.e.\ function free clauses with predicates of arity at most 2.
Moreover, \name avoids grounding for two logical rules: implication and equivalence relations.
As the second contribution, we prove that \name is fully expressive, i.e.\ for any ground truth of clauses of the above form, there exists a \name model (with embedding vectors) that represents that ground truth.
To the best of our knowledge, it is the first time that theoretical proofs are provided for the expressiveness of a neural network based embedding model. This proof indeed reassures us to inject different Horn rules in the model (encoding technique).
Regarding the encoding technique, our third contribution is that we additionally derive formulae for enforcing the model to learn different relations including (anti-)symmetric, implication, equivalence, inverse, transitive, composition, negation as well as irreflexive. To our knowledge, our model is the first model that can encode these rules as well as provides practical solution for encoding them. 

\section{Related Works}
\label{sec:RelatedWorks}

We investigate the related works in the light of two main issues we mentioned in the previous section, i.e.\ i) \textit{capability of a model} to encode rules and, ii) the \textit{encoding techniques}.
Moreover, we briefly review the relevant neural based models and show, in contrast to \name, they are not able to avoid grounding for the implication and equivalence relationships.

Considering the capability, \cite{SimplE} reports that TransE, FTransE, STransE, TransH and TransR have restrictions in encoding rules.
More concretely, TransE is incapable of encoding reflexive, symmetric and transitivity \cite{LogicTransX,TransH} and DistMult \cite{Distmult} cannot capture antisymmetric.
The CP decomposition cannot encode both symmetric and antisymmetric relations \cite{ComplexJourn}.
\citeauthor{ProofPaper} also investigate expressiveness of different bilinear models from a ranking perspective of their scoring matrix.

Despite the fact that score function of DistMult and ComplEx are similar, ComplEx can encode symmetric and antisymmetric relations due to the algebraic properties of complex numbers \cite{ComplEx}.
SimplE \cite{SimplE} is one the recent embedding model which is proven to be fully expressive. 
Moreover, conditions for encoding symmetric, antisymmetric and inverse patterns are derived. Although SimplE is fully expressive, for each entity/relation, two vectors should be provided which doubles the space. RotatE \cite{RotatE} is able to encode symmetric, antisymmetric and inverse and composition patterns. Although RotatE is shown to properly encode the patterns, a lot of negative samples should be generated together with a very big embedding dimension. It is indeed a big limitation when the model is trained on a large scale KG. 

Regarding encoding techniques, various approaches are introduced in the literature, which we review the most relevant ones.
As a preprocessing step, \cite{rocktaschel2015injecting}  iteratively infer new facts based on rules till no new facts can be inferred from a KG.
Then, they regard both ground atoms and existing rules as the set of new rules to be learned.
Accordingly, marginal probability of them 
are included in the training set and the loss function is minimized.
KALE \cite{guo2016jointly} uses margin ranking loss over logical formulae as well as triple facts and jointly learnins triples and formulae. 
In order not to rely on propositionalization for implication, \cite{demeester2016lifted} proposes a lifted rule injection method. 
\citeauthor{minervini2017regularizing} derive formulae for inverse and equivalence rules according to the score functions of TransE, ComplEx and DistMult. The obtained formulae are added to the objective as a regularization terms. 
Other methods that consider relation paths, which is closely connected to rules, are well-studied in the literature e.g.\ \cite{RelationPathneelakantan2015compositional,RelationPathlin2015modeling,Relationpathguu2015traversing}. 

There are also other ways of encoding rules, e.g.\
RUGE \cite{guo2018RUGE} presents a \textit{generic} (model-independent) framework to inject rules with confidence scores into an embedding model. 
The rules are encoded as constraints for an optimization problem. One of the main disadvantages of RUGE is that the model needs grounding of all rules. For example, to inject the rule $\forall h,t, \, if \,\, (h, \text{BornIn}, t) \xrightarrow{} (h, \text{Nationality}, t),$ $h,t$ should be replaced by all the entities that the triple $(h, \text{BornIn}, t)$ exists in the KG. 
In contrast to RUGE, \cite{SimpConstding2018improving} follows a \textit{model dependent} approach for injection of rules. It encodes non-negativity and entailment as constraints in ComplEx. 
It is shown \cite{SimpConstding2018improving} that the \textit{model dependent} approach of \cite{SimpConstding2018improving} outperforms the \textit{generic} approach of \cite{guo2018RUGE} on the FB15k dataset. However, \cite{SimpConstding2018improving} can only inject implication rule which is a limitation. 

As we mentioned, \name, in contrast to other relevant neural based models, avoids grounding for the implication and equivalence relationships.
E-MLP, ER-MLP, NTN, ConvE and ConvKB are among the most successful models in the literature \cite{socher2013reasoning,ER-MLPdong2014knowledge,socher2013reasoning,ConvEdettmers2018convolutional,ConvKBnguyen2017novel}.
The main common characteristics of all models is that $h$, $r$ and $t$ are treated as inputs or weights of  hidden layers while in \name $h$ and $t$ are inputs and $r$ is  the output of the network.
Having relations encoded as inputs or hidden layer weights requires that all groundings of the rules be fed into the network.
The detailed explanation of why \name is capable of avoiding grounding is properly addressed in Section~\ref{sec:OurMethod}.

To sum up, many models, like translation based models are incapable of encoding some rules.
The models which are reported to be capable, can either learn rules using existing triples in a KG or are enforced to learn by properly injecting the rules into their formulation.
The former kinds of models have still the risk of not properly learning rules as data in KGs are known to be very incomplete.
Therefore injecting rules enhance the learning performance of models.
Regarding fully expressiveness (FE), SimplE and  RESCAL, HolE etc are FE under some conditions \cite{ProofPaper}.

\begin{figure}[!h]
\centering
\includegraphics[height=3cm,width=6cm]{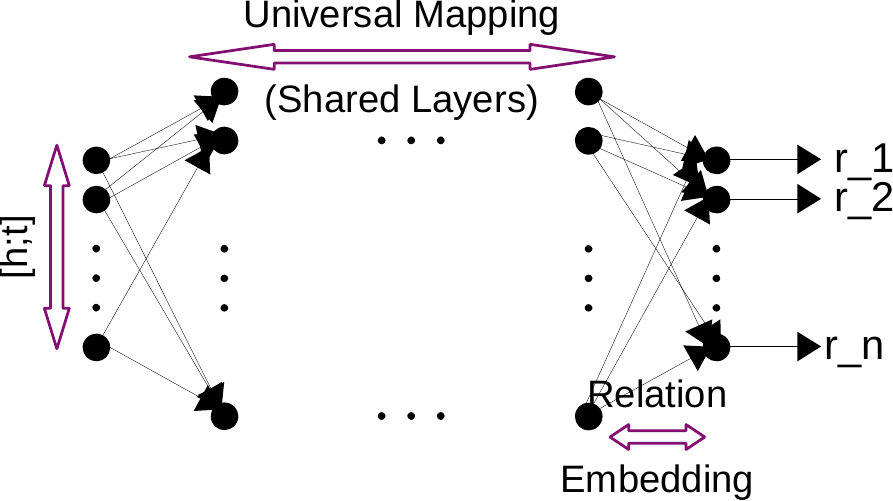}
\caption{\name: The hidden layer mapping, which is universal according to the theory of NN, is shared between entities and relations. One output node is associated to each relation.}
\label{fig:LogicENN}
\end{figure}

\section{The \name Approach}
\label{sec:OurMethod}

In this section we introduce our model and contribute to both the capability and the encoding technique.
We first present \name as a neural embedding model which is capable of encoding rules and we prove that it is fully expressive. 
We then discuss how we can algebraically formulate the rules and inject them into the model.
We then present our optimization approach in order to learn rules by \name.

This work considers clauses of the form ``$\text{premise}\Rightarrow\text{conclusion}$'', in which ``conclusion'' is an atom and ``premise'' is a conjunction of several atoms. 
Atoms are triples of type $(\var{x},r,\var{y})$ where $\var{x}, \var{y}$ are variables and ``$r$'' is a known relation in the KG.
We refer to such clauses as \textit{rules} from now on.

\subsection{The Proposed Neural Embedding Model}

It is known that using the same embedding space to represent both entities and relations is less competitive compared to considering two separate spaces~\cite{TransR}. 
This motivates to consider a neural network (NN) in which entities and relations are embedded in two different spaces.
Another motivation is that the previously reviewed NN approaches encode relations into the input layer or consider them as input weights of a hidden layer which is restrictive for avoiding grounding when one considers implication relationship.

We consider entity pairs as input and relations as output.
More precisely, we consider the embeddings of entity pairs,  $[\vec{h},\vec{t}]$, as input which together with weights are randomly initialized in the beginning.
During learning, \name optimizes both weights and embeddings of the entities according to its loss function.
The output weights of the network are embeddings of relations and the hidden layer weights are shared between all entities and relations as shown in Figure~\ref{fig:LogicENN}.
Despite \name takes embedding pairs of entities as input, it learns the embedding of each individual entity through a unique vector.
This is in contrast to some matrix factorization approaches which loose information by binding embedding vectors in form of entity-entity or entity-relations 
\cite{Nickel2015ReviewRelationalMLforKG}.

We denote the score function of a given triple $(h,r,t)$ by  $f^r(h,t)$, or more compactly by $f^r_{h,t}$.
Without loss of generality, we use a single hidden layer for the NN to show theoretical capabilities of \name and we define its score as:
\begin{align}\label{eq:NN}
    f^r_{h,t} = 
    \sum_{i=1}^L
    \phi(\langle \vec{w}_i , [\vec{h},\vec{t}] + b_i \rangle) \beta^r_i
    = \sum_{i=1}^L 
    \phi_{h,t}(\vec{w}_i,b_i) \beta^r_i  = \Phi_{h,t}^T \vec{\beta}^r
\end{align}
where $L$ is the number of nodes in hidden layer, $\vec{w}_i \in \mathbb{R}^{2d}$ and  $\beta^r_i \in \mathbb{R}$ are input and output weights of the $i$th hidden node respectively. 
$\vec{\beta}^r = [\beta_1^r, \ldots, \beta_L^r]^T$ are the output weights of the network which are actually embedding of relations.
That is because in the last layer a linear function acts as the activation function.
$\phi_{h,t}(\vec{w}_i,b_i) = \phi(\langle \vec{w}_i , [\vec{h},\vec{t}] + b_i \rangle)$ is the output of the $i$th hidden node and $\Phi_{h,t} = [\phi_{h,t}(\vec{w}_1,b_1), \ldots, \phi_{h,t}(\vec{w}_L, b_L)]^T$ is feature mapping of the hidden layer of the network which is shared between all relations. 
$\vec{h},\vec{t} \in \mathbb{R}^d$ are embedding vectors of head and tail respectively and $d$ is the embedding  dimension.
Therefore
$\vec{w}_i, [\vec{h},\vec{t}] \in  \mathbb{R}^{2d}$.
Finally, $\phi(.)$ is an activation function and $\langle.,.\rangle$ is the inner product.

\begin{table*}[!ht]
\centering
\resizebox{0.9\textwidth}{!}{

\begin{tabular}
{ p{1.7cm}p{3.5cm}p{3cm}p{4cm}p{5cm}}
 \toprule
  Rule 
  &  Definition \newline $\var{\forall} \var{h}, \var{t}, \var{s} \in \mathcal{E}: \ldots $
  & Formulation based on score function
  & Formulation based on NN
  & Equivalent regularization form \newline
  (Denoted as $\mathcal{R}_i$ in Equation~\eqref{eq:optimization} )\\
 \midrule 
 
  Equivalence   
  & $(\var{h},r_1,\var{t})\Leftrightarrow(\var{h},r_2,\var{t})$
  & $f_{h,t}^{r_1} = f_{h,t}^{r_2} + \xi_{h,t} $ & $\Phi_{h,t}^T (\vec{\beta}^{r_1}-\vec{\beta}^{r_2}) =  \xi_{h,t}$ 
  & $\max(\left\lVert\vec{\beta}^{r_1}-\vec{\beta}^{r_2}\right\lVert_{1}-\xi_{\text{Eq}},0)$\\
 \midrule

 Symmetric  
 & $(\var{h},r,\var{t})\Leftrightarrow(\var{t},r,\var{h})$
 & $f_{h,t}^r = f_{t,h}^r + \xi_{h,t} $
 & $(\Phi_{h,t}-\Phi_{t,h})^T \vec{\beta}^r = \xi_{h,t}$ 
 & $\max(|(\Phi_{h,t}-\Phi_{t,h})^T \vec{\beta}^r|-\xi_{\text{Sy}},0)$ \\
 \midrule

Asymmetric  
 & $(\var{h},r,\var{t})\Rightarrow\neg(\var{t},r,\var{h})$ 
 & $f_{h,t}^r = f_{t,h}^r + \mathcal{M}_{h,t} $
 & $(\Phi_{h,t}-\Phi_{t,h})^T \vec{\beta}^r = \mathcal{M}$ 
 & NC \\
 \midrule
 
 Negation 
 & $(\var{h},r_1,\var{t})\Leftrightarrow\neg(\var{h},r_2,\var{t})$ 
 & $f_{h,t}^{r_1} = \mathcal{M} - f_{h,t}^{r_2} + \xi_{h,t}$
 & $\Phi_{h,t}^T (\vec{\beta}^{r_1}+\vec{\beta}^{r_2}) = \mathcal{M} + \xi_{h,t}$ 
 & NC \\
 \midrule
 
 Implication  
 & $(\var{h},r_1,\var{t})\Rightarrow(\var{h},r_2,\var{t})$
 & $f_{h,t}^{r_1} \le f_{h,t}^{r_2}$ 
 & $\Phi_{h,t}^T (\vec{\beta}^{r_1}-\vec{\beta}^{r_2}) \leq 0 $ 
 & $\max(\sum_{i} (\vec{\beta}_i^{r_1} - \vec{\beta}_i^{r_2})+  \xi_{\text{Im}},0)$ \\
 \midrule
 
 Inverse
 & $(\var{h},r_1,\var{t})\Rightarrow(\var{t},r_2,\var{h})$
 & $f_{h,t}^{r_1} \le f_{t,h}^{r_2}$
 & $\Phi_{h,t}^T \vec{\beta}^{r_1}-\Phi_{t,h}^T\vec{\beta}^{r_2} \leq 0 $
 & $\max(\Phi_{h,t}^T \vec{\beta}^{r_1}-\Phi_{t,h}^T\vec{\beta}^{r_2}+\xi_{\text{In}},0) $\\
 \midrule
 
 Reflexivity 
 & $(\var{h},r,\var{h})$
 & $f_{h,h}^r = \mathcal{M} - \xi_{h,h}$
 & $\Phi_{h,h}^T {\vec{\beta}^r} = \mathcal{M} - \xi_{h,h}$ 
 & NC \\
 \midrule
 
 Irreflexive 
 & $\neg(\var{h},r,\var{h})$
 & $f_{h,h}^r = \xi_{h,h}$
 & $\Phi_{h,h}^T {\vec{\beta}^r} = \xi_{h,h}$ 
 & NC \\
 \midrule
 
 Transitivity    
 & $(\var{h},r,\var{t})\land(\var{t},r,\var{s}) \Rightarrow(\var{h},r,\var{s})$
 & $ \sigma(f^r_{h,s}) \geq \sigma(f^r_{h,t}) \times \sigma(f^r_{t,s})$ 
 & $\sigma(\Phi_{h,t} \vec{\beta}^r)
 \times \sigma(\Phi_{t,s} \vec{\beta}^r) - \sigma(\Phi_{h,s}^T {\vec{\beta}^r}) \leq 0 $ & $\max(\sigma(\Phi_{h,t} \vec{\beta}^r) \times \sigma(\Phi_{t,s} \vec{\beta}^r)- \sigma(\Phi_{h,s}^T {\vec{\beta}^r})+\xi_{\text{Tr}},0) $\\
\midrule

 Composition    
 & $(\var{h},r_1,\var{t})\land(\var{t},r_2,\var{s}) \Rightarrow(\var{h},r_3,\var{s})$
 & $\sigma(f^{r_1}_{h,s}) \geq \sigma(f^{r_2}_{h,t}) \times \sigma(f^{r_3}_{t,s})$ 
 & $\sigma(\Phi_{h,t} \vec{\beta}^{r_1}) \times \sigma(\Phi_{t,s} \vec{\beta}^{r_2})- \sigma(\Phi_{h,s}^T {\vec{\beta}^{r_3}}) \leq 0 $ 
 & $\max(\sigma(\Phi_{h,t} \vec{\beta}^{r_1}) \times \sigma(\Phi_{t,s} \vec{\beta}^{r_2})- \sigma(\Phi_{h,s}^T {\vec{\beta}^{r_3}})+\xi_{\text{Co}},0)  $ \\
\bottomrule
\end{tabular}

}
\caption{\footnotesize
Formulation and representation of rules (NC: Not considered for implementation).}
\label{tbl:LogicalConstraints}
\end{table*}

Due to having shared hidden layers in the design, \name is efficient in space complexity.
The space complexity of the proposed model is $\mathcal{O}(N_e d + N_r L)$ where $N_e, N_r$ are number of entities and relations respectively.

\subsection{Capability of the Proposed Network}

As mentioned in the section \ref{sec:Intro}, if a model is not fully expressive, it might be wrongly expected to learn a rule which is incapable of.  Therefore, investigation of the theories corresponding to the expressiveness of an embedding model is indeed important. 
Accordingly, we now prove that \name is fully expressive i.e., capable of representing every ground truth over entities and relations in a KG.  

Let $\mathcal{F}_L$ be the set of all possible neural networks with $L$ hidden nodes as defined by \eqref{eq:NN}.
Therefore, the set of all possible networks with arbitrary number of hidden nodes will be 
$\mathcal{F} = \bigcup_{l=1}^{\infty} \mathcal{F}_{l}$.
Let $\mathcal{C}(X)$ denote set of continuous functions over $\mathbb{R}$.
Let $\mathcal{E}$ be the set of entities and $e \in \mathcal{E}$ is an entity with embedding vector $\vec{e} \in \Omega_{\mathcal{E}} \subset \mathbb{R}^{d}$.
We also assume that $\Omega_{\mathcal{E}}$ is a compact set. We have the following theorem.

\begin{theorem}\label{theorem:expressiveness} 
Let $\mathcal{F}$ be the set of all possible networks defined as above, $\mathcal{F}$ be dense in $C(\mathbb{R}^{2d})$ where $d \geq 1$ is arbitrary embedding dimension.
Given any ground truth in a KG with $\tau$ true facts, there exists a \name in $\mathcal{F}$ with embedding dimension $d$, that can represent the ground truth. 
The same holds when $\mathcal{F}$ is dense in  $\mathcal{C}(\Omega)$ where $\Omega=\Omega_\mathcal{E}\times \Omega_\mathcal{E}$ is the Cartesian product of two compact sets.
\end{theorem}

The theorem proof as well as more detailed technical discussion are included in the supplementary materials of the paper.

\subsection{Formulating Rules}
\label{sec:FormulatingRules}

Let $a, b$ be two grounded atoms of a  clause as defined in the beginning of  Section~\ref{sec:OurMethod}, and let the truth values of $a$ and $b$ are denoted by $P(a)$ and $P(b)$ respectively. 
To model the truth values of negation, conjunction, disjunction and 
implication of $a$ and $b$ we define
$P(\neg a)$, $P(a \land b)$, $P(a \lor b)$ as in \cite{guo2018RUGE} but we define 
$P(a \Rightarrow b):  P(a) \leq P(b)$.
These can be used to derive formulation of rules based both on score function as well as the NN, as shown in Table~\ref{tbl:LogicalConstraints}.

As an example, consider the implication rule $\forall \var{h},\var{t}: (\var{h},r_1,\var{t})\Rightarrow(\var{h},r_2,\var{t})$.
Using $ P(a) \leq P(b)$, we can infer  
$f_{h,t}^{r_1} \le f_{h,t}^{r_2}$.
By \eqref{eq:NN}, we get $\Phi_{h,t}^T (\vec{\beta}^{r_1}-\vec{\beta}^{r_2}) \leq 0 $.
Provided that our activation function is positive, i.e.\ $\Phi_{h,t}^T \ge \vec{0}$, we will have $(\vec{\beta}^{r_1}-\vec{\beta}^{r_2}) \leq \vec{0} $.
That latter formula is independent of $h$ and $t$ which means we do not need any grounding for implication.
The same procedure shows that we can avoid grounding for equivalence. 
However for other rules this is not possible.

Using truth values defined as above, we can derive a formulation of a rule based on score function (e.g.\ $f_{h,t}^{r_1} \le f_{h,t}^{r_2}$ for implication) in the 3rd column of Table~\ref{tbl:LogicalConstraints}, and its equivalent formulation based on the NN of \eqref{eq:NN} in the 4th column.
Assume $\mathcal{M}> 0$ indicates True and $0$ indicates False.
We now state the necessity and sufficiency conditions for \name to infer various rules.
For the proof we can do similar procedure to the one we did for  implication.
The detailed proof is provided in the supplementary materials of the paper.

\begin{theorem}\label{theorem:infer}
For all $h,t,s \in \mathcal{E}$, set $\xi_{h,t}=0$ in column ``Formulation based on NN'' of Table \ref{tbl:LogicalConstraints}.
For each relation type given in Table \ref{tbl:LogicalConstraints}, \name can infer it if and only if the corresponding equation in column ``Formulation based on NN'' holds ($\xi_{h,t}$ is set to 0).
\end{theorem}

Since KGs may contain wrong data or facts with less truth confidence \cite{SimpConstding2018improving}, the assumption of $\xi_{h,t}=0$ in Theorem \ref{theorem:infer} is too rigid in practice. 
Therefore, as shown in Table \ref{tbl:LogicalConstraints} we consider $\xi_{h,t}$ to be a slack variable that allows enough flexibility to deal with uncertainty in KG. 
This allows us to infer uncertainty as $\xi$ through a validation step of learning.
Although considering $\xi_{h,t}$ as slack variables improves flexibility, due to grounding we will have too many slack variables.
Therefore, in the implementation level we decided to consider one slack variable for each relation type
e.g.\ one $\xi_{\text{Eq}}$ was used for all the equivalence relations (see the last column of Table~\ref{tbl:LogicalConstraints}). This enables the model to mitigate the negative effect of uncertainty of rules by considering average uncertainty per rule type. Experimental results show the effectiveness of inclusion of a slack variable per a rule type. During experiments, we obtained the hyper-parameters corresponding to each rule type sequentially through validation step. Therefore, instead of having $n_1 \times n_2 \times \ldots \times n_m$ combinations for hyper-parameter search corresponding to the rules injection, we have $n_1 + n_2 + \ldots + n_m, $combinations where $n_i$ refers to the number of candidates for the slack variable of the $i$-th rule type. Experimental results confirms that this approach gets satisfactory performance as well as significant reduction in the search space.

\subsection{Rule Injection and Optimization}
To inject rules into embeddings and weights of \eqref{eq:NN}, we define the following optimization:

\begin{equation}\label{eq:optimization}
\begin{aligned}
&\min_{\theta} \, \sum_{(h,r,t)\in \mathcal{S} } 
&&\alpha_{h,t}^r\log(1+\exp(-y_{h,t}^r\, f^r_{h,t})) + \lambda  \sum_{i=1}^l  \frac{\mathcal{R}_i}{N_i}  \\
&\text{subject to} &&\|h\| = 1 \; \text{and} \; \|t\| = 1  \,.
\end{aligned}
\end{equation}

where $\mathcal{S}$ is the set of all positive or negative samples, $\alpha_{h,t}^r$ is set to 1 for positive samples.
For negative samples if we get big scores, then the model should suppress it by enforcing a big value for $\alpha_{h,t}^r$ using formula No.5 in \cite{RotatE}.
$\mathcal{R}_i$ refers to the $i$th rule, $N_i$ is the number of groundings, $\lambda$ is a regularization term and $y_{h,t}^r$ represents the label of $(h,r,t)$ which is 1 for positive and -1 for negative samples.

In \eqref{eq:optimization}, we use negative log-likelihood loss with regularizations over logical rules.
Loss as the first term focuses on learning facts in KG while the second term, i.e.\ regularization, injects rules into the learning process.
The regularization are provided as penalties in the last column of Table~\ref{tbl:LogicalConstraints}.

\section{Experiments and Discussions}
\label{sec:Experiments}

\begin{table*}[!ht]
    \footnotesize
    
    \centering
    \renewcommand\tabcolsep{2.0pt}
    \resizebox{0.9 \textwidth}{!}{
      \begin{tabular}{lcccccccccccc}  
        \toprule  
        
        &\multicolumn{3}{c}{FB15k -- Raw}&\multicolumn{3}{c}{FB15k -- Filtered}&\multicolumn{3}{c}{WN18 -- Raw}&\multicolumn{3}{c}{WN18 -- Filtered}\cr 
            \cmidrule(lr){2-4} \cmidrule(lr){5-7}\cmidrule(lr){8-10} \cmidrule(lr){11-13}
            &MR&Hits@10&MRR&FMR&FHits@10&FMRR&MR&Hits@10&MRR&FMR&FHits@10&FMRR\cr
        \midrule
        TransE \cite{bordes2013translating} &201&43.4&18.4&70&61.8&30.7&263&75.4&-&251&89.2&-\cr
        DistMult \cite{Distmult} &279&50.0&25.5&120.4&84.2&70.5&-&-&-&655&94.6&79.7\cr
        ComplEx \cite{ComplEx} &266&48.5&23.0&106&82.6&67.5&573&82.1&58.7&543&94.7&94.1\cr
        ANALOGY \cite{liu2017analogical} &279&50.5&26.0&121&84.3&72.2&-&-&65.7&-&94.7&94.2\cr
        ConvE \cite{ConvEdettmers2018convolutional} &191&52.5&27.2&51&85.1&68.9&-&-&-&504&\textbf{95.5}&94.2\cr
        SimplE \cite{SimplE} &-&-&24.2&-&83.8&72.7&-&-&58.8&-&94.7&94.2\cr
        RotatE \cite{RotatE} &162&57.5&31.0&74&80.6&61.8&636&84.2&66.2&627&94.6&93.0\cr
        QuatE \cite{zhang2019quaternion}
        &182&52.6&27.0&\textbf{37}&79.1&56.1&402&81.9&58.0&386&95.7&92.8\cr
        \midrule
        PTransE \cite{RelationPathlin2015modeling} &207&51.4&-&58&84.2&-&-&-&-&-&-&-\cr
        KALE \cite{guo2016jointly} &225&47.5&21.3&73&76.2&52.3&\textbf{252}&83.3&39.5&\textbf{241}&94.4&53.2\cr
        RUGE \cite{guo2018RUGE} &203&55.3&28.5&97&86.5&76.8&-&-&-&-&-&-\cr
        ComplEx-NNE+AER \cite{SimpConstding2018improving} &193&57.3&29.3&116&\textbf{87.4}&\textbf{80.3}&481&83.5&61.9&450&94.8&\textbf{94.3}\cr
        \midrule
        LogicENN$_R^*$ (our work)&\textbf{175}&\textbf{66.9}&\textbf{40.2}&112&\textbf{87.4}&76.6&368&\textbf{84.2}&\textbf{66.3}&357&94.8&92.3\cr
        \bottomrule
    
    \end{tabular}       
    } 
    \caption{
    Link prediction results. 
    Rows 1-8: basic models with no rules.
    Rows 9-12: models which encode rules. 
    Results on FB15k in rows 1-5 are taken from \protect\cite{akrami2018reevaluation}
    while the rest are taken from original papers/code.
    Dashes: results could not be obtained.
    }
    \label{tbl:AllResults}
\end{table*}



To show the capability of \name, we evaluated it on the link prediction task.
The task is to complete a triple $(h,r,t)$ when $h$ or $t$ missing, i.e.\ to predict $h$ given $(r,t)$ or $t$ given $(h,r)$.
For evaluation, we will use Mean Rank (MR), Mean Reciprocal Rank (MRR) and Hit@10 in the raw and filtered settings as reported in \cite{wang2017surveyEmbeddings,TransR}.

\paragraph{Datasets.}
We used FB15k and WN18 with the settings reported in \cite{bordes2013translating}.
We used the rules reported in \cite{guo2018RUGE} for FB15k, and the rules in \cite{guo2016jointly} for WN18.
The confidence levels of rules are supposed to be no less than 0.8. 
Totally we used 454 rules for FB15k and 14 rules for WN18.
As both data sets are reported \cite{ConvEdettmers2018convolutional}
to have the inverse of triples in their test sets, it is argued that the increase of performance of some models might be due to the fact that models have learned more from the inverses rather the graph itself.
Using these data sets to compare just the rule-based models would be fine as all are using rules, e.g.\ experiments of RUGE \cite{guo2018RUGE}.
However when one wants to compare rule-based with non-rule-based models, it would be better to use a data set which has not much inverses.
As FB15k-237 has already addressed this, we used it to compare \name with other non-rule-based models.

To formulate rules, we categorized them according to their definitions in Table~\ref{tbl:LogicalConstraints}. 
We did grounding for all rules except for those denoted by NC, as well as equivalence and implication since \name does not need them by formulation (see Sec.~\ref{sec:OurMethod}). 
To maximize the utility of inferencing, like RUGE, we take as valid groundings whose premise triples are observed in the training set, while conclusion triples are not.

\begin{table}[!ht]
\footnotesize

\resizebox{\columnwidth}{!}{

\begin{tabular}{lcccccc}
\toprule
 & \multicolumn{2}{c}{MR} & \multicolumn{2}{c}{Hit@10} & \multicolumn{2}{c}{MRR} \\
\cmidrule{2-7} 
 & ReLU & $\sigma$       
 & ReLU & $\sigma$
 & ReLU & $\sigma$   \\
 \midrule
 With No Rule
 & 320 & 430 & 42.2 & 36.6 & 18.1 & 15.2  \\
 
Inverse$^\star$
 & \textbf{187} & \textbf{180} & \textbf{62.1} & \textbf{59.3} & \textbf{37.7} & \textbf{35.2} \\
 
Implication
 & 321 & 421 & 40.5 & 37.1 & 18.2 & 16.4 \\
 
 Symmetry
 & 299 & 387 & 42.2 & 37.9 & 18.5 & 17.2 \\
 
 Equivalence
 & 302 & 330 & 41.7 & 38.4 & 19 & 17.7 \\
 
 Composition
 & 303 & 391 & 41 & 37.1 & 18.1 & 16 \\    
 \bottomrule
\end{tabular}

} 

\caption{Link prediction using different activation functions and rules (FB15k).
``ReLU'', ``$\sigma$'' respectively correspond to LogicENN$_{R}$, LogicENN$_{S}$.
$\star$: the model outperformed all baselines.}

\label{tbl:ResultsWithRules}

\end{table}

\begin{table}[]
    \centering
    \footnotesize
    \begin{tabular}{lcccc}
        \toprule
        &\multicolumn{2}{c}{Raw} &\multicolumn{2}{c}{Filtered}\cr
        \cmidrule(lr){2-3} \cmidrule(lr){4-5}
        &MR&Hits@10 &MR&Hits@10\cr
        \midrule
        ComplEx & 620 & 25.4 &457&45.7\cr
        ComplEx-N3 & 553 & 29.7& 421&50.0\cr
        ConvE & 489 & 28.4&246&49.1 \cr
        ASR-COMPLEX & 570 & 26.3&420&46.1 \cr
        RotatE & 374 & 33.3&258&47.1 \cr
        QuatE & \textbf{354} & 32.2&161&48.3 \cr
        \midrule
        LogicENN$_{R}^*$ & 454 & \textbf{34.7}&424&47.3 \cr
        \bottomrule
    \end{tabular}
    \caption{Link prediction results on FB15k-237.}
    \label{tbl:Results_FB237}
\end{table}

\paragraph{Experimental Setup.}
To select the structure of the model, we tried different settings for the number of neurons/layers and types of activation functions.
Two of the best settings were LogicENN$_{R}$ and LogicENN$_{S}$ which both had 3 hidden layers with 1k, 2k and 200 neurons respectively. 
The 4th layer was the output layer where the number of neurons were equal to the number of relationships in the datasets.

In LogicENN$_{R}$ we used ReLU on all hidden layers and in LogicENN$_{S}$ we used Sigmoid as activation functions between layers and ReLU on the last hidden layer.
Both of LogicENN$_{R}$ and LogicENN$_{S}$ were combined with each of rules we used.
When models were armed with all existing rules for a dataset, we denote them by LogicENN$_{R}^*$ and LogicENN$_{S}^*$ respectively.
Moreover, LogicENN$_{RR}^*$ denotes our approach when we have added the reverse of triples in the target data set, as also done in \cite{lacroix2018canonical}.

We implemented the models in PyTorch and used the Adam optimizer for training. 
We select the optimal hyperparameters of our models by early validation stopping according to MRR on the validation set. We restricted the iterations to 2000.
For basic models of LogicENN$_{R}$ and LogicENN$_{S}$ which integrate no rules, we created 100 mini-batches on each dataset.
We tuned the embedding dimensionality $d$ in \{$100, 150, 200$\}, the learning rate $\gamma$ in \{$0.0001, 0.0005, 0.001, 0.005, 0.01$\} and the ratio of negatives over positive training samples $\alpha$ in \{$1, 2, 3, 5, 8, 10$\}. 
The optimal configuration for both LogicENN$_{R}$ and LogicENN$_{S}$ are: $d$ = 200, $\gamma$ = 0.001, $\alpha$ = 8 on FB15k; and $d$ = 200, $\gamma$ = 0.001, $\alpha$ = 5 on WN18. 
Based on LogicENN$_{R}$ and LogicENN$_{S}$ with their optimal configurations, we further tuned the regularization coefficient $\lambda$ in \{$0.001, 0.005, 0.01, 0.05, 0.1, 0.5, 1 $\} and slack variables $\xi_i$ in \{$0, 0.1, 0.5, 1, 3, 5, 10$\} for different types of rules (see Table~\ref{tbl:LogicalConstraints}) to obtain all optimal hyperparameters of LogicENN$_{R}^*$ and LogicENN$_{S}^*$ which integrate all rules in datasets.
For LogicENN$_{R}^*$, we find the following hyperparameters are optimal: $\lambda$ = 0.05, $\xi_{\text{Eq}}$ = 1, $\xi_{\text{Sy}}$ = 0.5, $\xi_{\text{Im}}$ = 5, $\xi_{\text{Co}}$ = 0.1, $\xi_{\text{In}}$ = 3 on FB15k; $\lambda$ = 0.01, $\xi_{\text{In}}$ = 0.1 on WN18.
The optimal configurations of LogicENN$_{S}^*$ on FB15k are : $\lambda$ = 0.05, $\xi_{\text{Eq}}$ = 0.5, $\xi_{\text{Sy}}$ = 0.1, $\xi_{\text{Im}}$ = 3, $\xi_{\text{Co}}$ = 0.1, $\xi_{\text{In}}$ = 3.
For WN18 they are $\lambda$ = 0.005, $\xi_{\text{In}}$ = 0.1.

\paragraph{Results.}
Table~\ref{tbl:AllResults} shows comparison of \name with eight state-of-the-art embedding models as basic models which only use observed triples in KG and rely on no rules. 
We also took PTransE, KALE, RUGE, ComplEx-NNE+AER as additional baselines.
They encode relation paths or logical rules like LogicENN$^*_{R/S}$. 
Among them, the first two are extension of TransE, while the rest are extensions of ComplEx.

To compare both raw and filtered results of LogicENN$^*_{R/S}$ and  baselines, we take the results of the first five baseline models on FB15k reported by \cite{akrami2018reevaluation} and use the code provided by \cite{guo2016jointly,guo2018RUGE,SimpConstding2018improving} for KALE, ComplEx, RUGE and ComplEx-NNE+AER to produce the raw results with the optimal configurations reported in the original papers. We also ran the code of RotatE to get its results. Because RotatE \cite{RotatE} uses Complex vectors, we set its embedding dimension to 130 (260 adjustable parameters per each entity) and generate 10 negative samples to have a fair comparison to our method. Results of QuatE \cite{zhang2019quaternion} are obtained by running their code with embedding dimension 60 and 10 negative samples without using type constraint to have a fair comparison to our method. We set the embedding dimension to 60 (240 adjustable parameters) because QuatE provides 4 vectors for each entity.
Other results in Table~\ref{tbl:AllResults} are taken from the original papers.

As we previously discussed, FB15k and WN18 have the inverse of triples in their test sets. 
To show the performance of \name in comparison of non-rule-based methods we ran experiments on FB15k-237 which is reported to have not much reverse of triples.
Table~\ref{tbl:Results_FB237} shows the comparison of our method with other non-rule-based models in this regard.

\paragraph{Discussion of Results.}
As shown in Table~\ref{tbl:AllResults},
LogicENN$^*_{R}$ outperformed all embedding models on FB15k in the raw setting using MR, Hit@10 and MRR.
For the filtered setting it also performs better considering FHit@10 and very close to RUGE (the 2nd best performing model) using FMRR. 
Considering WN18, our model got the best performance in Raw-Hit@10 and Raw-MRR. 
In the terms of FHit@10, only ConvE  and QuatE outperformed our model.

To investigate whether inclusion of logical rules improve the performance of our model, we added each rule to the naked \name separately. 
Table~\ref{tbl:ResultsWithRules} shows the improvements by adding each rules separately. 
As shown, inclusion of each rule improves the performance of the naked model. 
For FB15k, the best improvement is obtained by the inverse rule which is the most common rule in FB15k.
Two variants of model i.e.\ LogicENN$_S$ and LogicENN$_R$ performed better when rules added.

The two most recent methods of RUGE and ComplEx-NNE+AER use ComplEx score function to encode rules.
As Table~\ref{tbl:AllResults} shows, the performance of ComplEx on Raw-Hit@10 was 48.5\% which encoding of rules by RUGE and ComplEx-NNE+AER improved it by less than 10\% (to 55.5\% and 57.3\% respectively).
In contrast, our method without any rule-encoding performed around 40\% (Table~\ref{tbl:ResultsWithRules}) which jumped to around 67\% (Table~\ref{tbl:AllResults}) when rules were encoded.
It shows our method improved around 27\% which is more than double of its competitors.
Therefore we can conclude that the encoding techniques of Table~\ref{tbl:LogicalConstraints}  can properly encode rules.

\begin{figure}
    \centering
    \includegraphics[height=4cm,width=7cm]{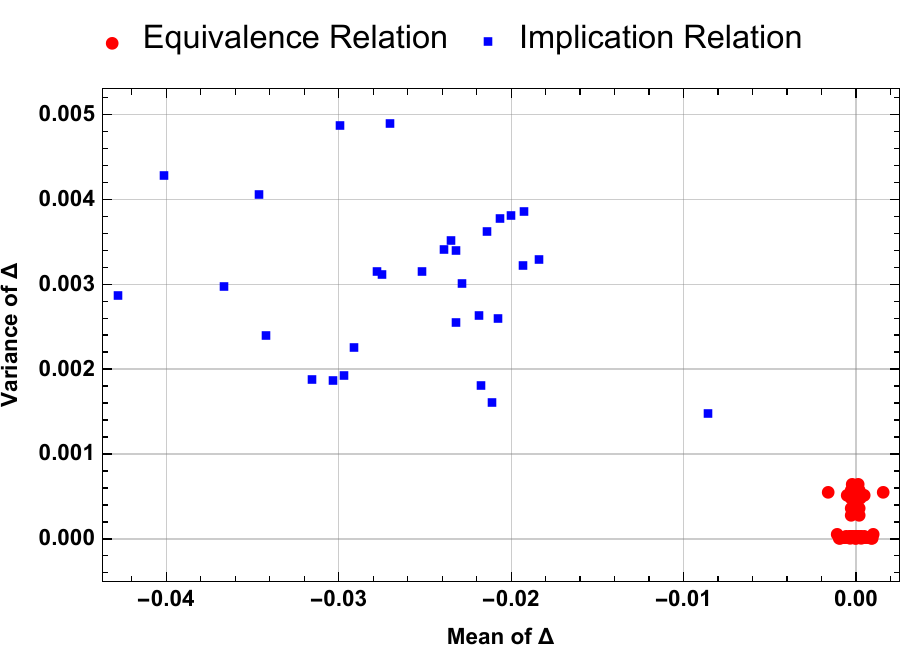}
    \caption{Statistics of difference vectors of equivalence relations ($\Delta_{\text{Equivalence}}$) and implication relations ($\Delta_{\text{Implication}}$).}
    \label{fig:stat}
\end{figure}

Figure~\ref{fig:stat} shows that the model has properly learned the equivalence and implication relations.
To better comprehend that, in Section~\ref{sec:FormulatingRules} we already explained that we can avoid grounding for the implication rule and the resulting formula was $\Delta_{\text{Implication}}:=(\vec{\beta}^{r_1}-\vec{\beta}^{r_2})  $ and we had $\Delta_{\text{Implication}} \leq \vec{0}$.
Similar argument implies that 
$\Delta_{\text{Equivalence}}=\vec{0}$.
Therefore, if the model has properly learned implication (equivalence) the differences of the embedding vectors of these two relations should contain negative (zero) elements.
In Figure~\ref{fig:stat}, the x-axis represents the means of the elements of the two $\Delta$s and the y-axis represents their variances. 
As depicted, the points associated to the equivalence relations are accumulated around the origin and the points associated to the implication relations are negative.
This shows that \name has properly encoded these rules without using grounding for 30 implication and 68 equivalence relations in FB15k.

As Table~\ref{tbl:Results_FB237} show,
LogicENN$^*_{R}$ outperforms all state-of-the-art in the terms of Raw Hits@10. 
We should note that the originally reported result of ComplEx-N3 used $d$=2k as the embedding dimension \cite{lacroix2018canonical}.
To have a fair and equal comparison, we re-ran their code with the same setting we used for all of our experiments, i.e.\ we used embedding dimension of 200.

\section{Conclusion and Future Work}
\label{sec:Conclusion}

In this work we introduced a new neural embedding model (\name) which is able to encode different rules.
We proved that \name is fully expressive and we derived algebraic formulae to enforce it to learn different relations.
We also showed how rules can be properly injected into learning.

Our extensive experiments on different benchmarks show that \name
outperformed all embedding models on FB15k in the raw and performed very well in the filtered setting.
For WN18, the model performed better than almost all others in the raw and very close to the best models in the filtered settings.
On FB15k-237 the model was better than non-rule-based models on raw Hit@10.

The expressiveness of other kinds of neural models as well as the necessity and sufficiency conditions for injecting rules, are targets of future work.

\clearpage

\bibliographystyle{abbrv} 
\bibliography{ijcai19.bbl}



\section{Supplementary Materials for the Paper: \name}

This section contains supplementary materials for our paper called: ``\textit{LogicENN: A Neural Based Knowledge Graphs Embedding Model with Logical Rules}''

In Section~\ref{sec:relevantNNs}, we first review the relevant neural based models and describe their similarities and differences to \name.
Then in Section~\ref{sec:Proofs}, we will provide our proposed theorems with their detailed proofs.

\subsection{LogicENN vs State-of-the-art Neural Based Models}
\label{sec:relevantNNs}

This section describes the relevant neural network based embedding models.
We divided them to models which do not consider logical rules into their embeddings and those which consider rules.
We then compare their score functions with LogicENN and discuss how it is different from other state-of-the-art models.

Before progressing more, we first define the relevant notations.
Vectors of are denoted by bold non-capital letters (e.g.\ $\vec{h}$ or $\vec{0},\vec{1}$ as vectors of zeros and ones respectively), matrices by bold capital letters and tensors by underlined bold capital letters.
$\mat{W}^{(.)}$ are weight matrices which have no dependency on $h$, $r$ or $t$, while $\mat{W}^{(.)}_{r}$ ($\ten{W}_{r}$) means that the weight matrix (tensor) is associated to relation $r$.
Moreover, $\Bar{\vec{h}}, \Bar{\vec{r}}$ and $\Bar{\vec{t}}$ are 2D-reshape of $h$, $r$ and $t$ respectively.

\subsubsection{Neural Network Based Embedding Models}

\paragraph{E-MLP.} \cite{socher2013reasoning} which is standard Multi Layer Perceptron (MLP) for KGE. E-MLP uses one neural network per each relations in the KG which has high space complexity. 

\paragraph{ER-MLP.} \cite{ER-MLPdong2014knowledge} In contrast to E-MLP which uses one neural network per each relation, ER-MLP shares its weights between all entities and relations. The relation embedding is placed as input of the network. 

\paragraph{NTN.} \cite{socher2013reasoning} employs 3-way tensor in the hidden layer to better capture interactions between features of two entities. The score function of NTN is as follows:
\begin{align}
    f^r_{u,v} = \boldsymbol{w}_{1r}^T \tanh(\boldsymbol{u}^T \boldsymbol{W}_{0htr}\boldsymbol{v} + W_{0hr} \boldsymbol{u} + W_{0tr} \boldsymbol{v} + b_r)
\end{align}

where $\boldsymbol{W}_{0htr} \in \mathbb{R}^{d\times d\times L}, b_r \in \mathbb{R}^L$ are the 3-way relation specific tensor and bias of hidden layer respectively. 

\paragraph{ConvE.} ~\cite{ConvEdettmers2018convolutional} is a multi-layer convolutional network for link prediction. The score function of ConvE is as follows:
\begin{align}
    f^r_{u,v} = g(\text{Vec}(g([\Bar{\boldsymbol{u}};\Bar{\boldsymbol{r}}]*\omega))W)\boldsymbol{v}
\end{align}
where $\omega$ is filter and $W$ is a linear transformation matrix. $g$ is an activation function.

\paragraph{ConvKB.} ~\cite{ConvKBnguyen2017novel} is a multi-layer convolutional network for link prediction with the following score function:

\begin{align}
    f^r_{u,v} = \text{Concat}(g([\Bar{\boldsymbol{u}};\Bar{\boldsymbol{r}},\Bar{\boldsymbol{v}}]*\omega)) W
\end{align}

\begin{table}[!h]
    \centering
    \footnotesize

\begin{tabular}{ll}
\toprule 
Method  & Score Function ($f^r_{h,t})$\\
\midrule 

E-MLP  & $ \vec{w}_{r}^T \tanh(\mat{W}^{(1)}_{r} \vec{h} + \mat{W}^{(2)}_{r} \vec{t})$ \\

ER-MLP  & $\vec{w}^T \tanh(\mat{W}^{(1)} \vec{h} + \mat{W}^{(2)} \vec{r} + \mat{W}^{(3)} \vec{t})$ \\

NTN  & $\vec{w}_{r}^T \tanh(\vec{h}^T \ten{W}_{r} \vec{t} + \mat{W}^{(1)}_{r} \vec{h} + \mat{W}^{(2)}_{r} \vec{t} + \vec{b}_r)$ \\

ConvE  & $g( \text{Vec}(g([\Bar{\vec{h}};\Bar{\vec{r}}]*\omega)) \, \mat{W} ) \, \vec{t}$ \\
ConvKB  & $\text{Concat}(g([\Bar{\vec{h}};\Bar{\vec{r}},\Bar{\vec{t}}]*\omega))\, \vec{w}$ \\
\bottomrule 
\end{tabular}

\caption{Comparison of score functions of the state-of-the-art relevant models.}

\label{tbl:NNEs}

\end{table}


\paragraph{SENN.} ~\cite{SENNguan2018shared} defines three multi-layer neural networks with Relu activation function for head, relation and tail prediction. Then, it integrates them into one loss function to train the model.

\paragraph{TBNN.} ~\cite{TBNNhan2018triple} is a triple branch neural network in which parallel branched layers are defined on the top of an interaction layer where each embedding of any element of a KG is specified by its multi restriction. The loss function is defined based on the score of three elements of each triple.

\paragraph{ProjE.} ~\cite{projeshi2017} is a two layer neural network. The first layer is a combination layer which works on tail and relation and the second layer is a projection layer which projects the obtained vector from the last layer to the candidate-entity matrix. The candidate-entity matrix is a subset of entity matrix where entities can be sampled in different ways. 

In short, Table~\ref{tbl:NNEs} specifies the score functions of different neural based embedding models. 

\subsubsection{KG Embedding Models with Logical Rules}

\paragraph{RUGE.}
provides a general framework to iteratively inject logical rules in KGE. Given a set of soft logical rules $\mathcal{R} = \{(l_p, \lambda_p)\}$ where $l_p$ is a rule and $l_p$ is its confidence value, rules are represented as mathematical  constraints to obtain soft labels for unlabeled triples. Then, an optimization problem is solved to update embedding vectors based on hard and soft labeled triples. The framework is used to train ComplEx model as case study. 

\paragraph{ComplEx-NNE+AER.} ~\cite{SimpConstding2018improving}, which is a model dependent approach, derives formula for entailment rule to avoid grounding in ComplEx. The model outperforms RUGE on FB15K in the terms of Meanrank and Hit@k.

\subsubsection{Comparison of LogicENN with Other Models}

LogicENN uses the scoring function (1). 
We formulate the score function to separate the entity and relation spaces.
This enables the model to map entities by a universal hidden layer mapping $\Phi_{h,t}$. 
Since we prove that $\Phi_{h,t}$ is universal, we can share it between all relations. 
Since $\Phi_{h,t}$ is used by several relations, we have fewer parameters. 
Several neural embedding models such as NTN and E-MLP didn't share parameters of hidden layer. 
Therefore, for each relation, a separate neural network is used.
ER-MLP feeds entity pairs as well as relation to the neural network.
Inclusion of relation in the hidden layer disables the model to avoid grounding for implication rule. 
The same problem happens for ConvE and ConvKB. 
Moreover, full expressiveness of ConvE and ConvKB is not investigated yet.  

Regarding encoding techniques, we derive formula for the proposed NN to encode function free Horn clause rules. 
For implication and equivalence rules, we approximate the original formula by avoiding grounding. 
Since we proved that our model is fully expressive, we can encode all Horn clause rules. 

Regarding the last column of the Table 1, we add slack variables to better handle uncertainty during injection of rules in the embeddings. 
The uncertainty is inherited from the fact that KG contain False positive triples.

\subsection{Theorems and Proofs}
\label{sec:Proofs}

In this section we state the theorems which prove the full expressiveness of our proposed model \name.

\begin{theorem}\label{theorem:expressiveness} 
Let $\mathcal{F}$ be the set of all possible networks defined as above, $\mathcal{F}$ be dense in $C(\mathbb{R}^{2d})$ where $d \geq 1$ is arbitrary embedding dimension.
Given any ground truth in a KG with $\tau$ true facts, there exists a \name in $\mathcal{F}$ with embedding dimension $d$, that can represent the ground truth. 
The same holds when $\mathcal{F}$ is dense in  $\mathcal{C}(\Omega)$ where $\Omega=\Omega_\mathcal{E}\times \Omega_\mathcal{E}$ is the Cartesian product of two compact sets.
\end{theorem}

\begin{proof}
Regarding the assumption of the theorem, $\Omega_{\mathcal{E}}$ is a compact set.
$\Omega$ is a compact set, since the Cartesian product of two compact set is also compact \cite{kuttler2011multivariable}.
Regarding lemma 2.1 in \cite{huang2000classification}, given $N$ disjoint regions $K_1, \ldots, K_{N} \subset \mathbb{R}^{2d}$, there exists at least one continuous function $f: \mathbb{R}^{2d} \rightarrow \mathbb{R}$ 
such that 
$f(x) = c_i$ when $x \in K_i, \, i=1, \ldots, N$.
$c_i, \, i=1,\ldots,N$ are arbitrary distinct constant values. 
Therefore, dealing with ground truth, $c_i \in \{0,1\}$, there exists a continuous function $f$ that represents the ground truth. 
Because $\mathcal{F}$ is dense in $C(\mathbb{R}^{2d})$ or $C(\Omega)$, there exists at least one neural network in  $\mathcal{F}$ that approximates the function $f$. 
As a conclusion, there exists a LogicENN that can represent the ground truth. 
\end{proof}

\textbf{Remark:} The density assumption of $\mathcal{F}$ in Theorem \ref{theorem:expressiveness} depends on the activation function of Equation (1) of the paper.
When the activation function is continuous, bounded, and non-constant, then $\mathcal{F}$ is dense in $\mathcal{C}(\Omega)$ for every compact set $\Omega$.
When it is unbounded and non-constant, then the set is dense in $\mathcal{L}^p(\mu)$  for all finite measure $\mu$. 
In this case, the compactness condition can be removed.
For non-polynomial activation functions which are locally essentially bounded, the set is dense in $\mathcal{C}(\mathbb{R}^d)$.

\begin{theorem}\label{theorem:infer}
For all $h,t,s \in \mathcal{E}$, set $\xi_{h,t}=0$ in column ``Formulation based on NN'' of Table \ref{tbl:LogicalConstraints}.
For each relation type given in Table \ref{tbl:LogicalConstraints}, \name can infer it if and only if the corresponding equation in column ``Formulation based on NN'' holds ($\xi_{h,t}$ is set to 0).
\end{theorem}

\begin{proof}[Proof for the Equivalence Relation.]
Based on the theorem statement, we want to show that \name can infer equivalence rule if and only if 
$\Phi_{h,t}^T (\vec{\beta}^{r_1}-\vec{\beta}^{r_2}) = 0.$
 
If r is an equivalence relation, we have:
$$ (h,r_1, t) \leftrightarrow (h,r_2,t). $$

Without loss of generality, let $\mathcal{M}, 0$ show scores of NN for positive and negative triples respectively. To be equivalence, both triples should be true or false simultaneously. Therefore, 
if $f_{r_1}(h,t) = \mathcal{M}$ then
$f_{r_2}(h,t) = \mathcal{M}$ and if $f_{r_1}(h,t) = 0$ then
$f_{r_2}(h,t) = 0$. 
We conclude that $f_{r_1}(h,t) = f_{r_2}(h,t)$. 

From Equation (1) in the paper, we have
$\Phi_{h,t}^T \vec{\beta}^{r_1}- \Phi_{h,t}^T \vec{\beta}^{r_2} = 0.$
Therefore, we have
$\Phi_{h,t}^T (\vec{\beta}^{r_1}-\vec{\beta}^{r_2}) = 0.$

\end{proof}

\begin{proof}[Proof for the Symmetric Relation.]

Based on the theorem statement, we want to show that \name can infer symmetric rule if and only if 
$(\Phi_{h,t}^T - \Phi_{t,h}^T) \vec{\beta}^{r} = 0$.
 
If r is an Symmetric relation, we have 
$$ (h,r, t) \leftrightarrow (t,r,h). $$

To be Symmetric relation, both triples should be true or false simultaneously. Therefore, 
if $f_{r}(h,t) = \mathcal{M}$ then
$f_{r}(t,h) = \mathcal{M}$ or if $f_{r}(h,t) = 0$ then
$f_{r}(t,h) = 0$. 
We conclude 
$f_{r}(h,t) = f_{r}(t,h)$.

From Equation (1) in the paper, we have
$\Phi_{h,t}^T \vec{\beta}^{r_1}- \Phi_{h,t}^T \vec{\beta}^{r_2} = 0.$

Therefore, we have:
$$\Phi_{h,t}^T \vec{\beta}^{r}= \Phi_{t,h}^T \vec{\beta}^{r}$$

We conclude that:
$$(\Phi_{h,t}^T \vec{\beta}^{r} - \Phi_{t,h}^T) \vec{\beta}^{r} = 0.$$

\end{proof}

\begin{proof}[Proof for the Implication Relation.]

Based on the theorem statement, we want to show that \name can infer implication rule if and only if 
$\Phi_{h,t}^T (\vec{\beta}^{r_1}-\vec{\beta}^{r_2}) \leq 0$. 
 
If r is implication rule, we have:
$$ (h,r_1, t) \rightarrow (h,r_2,t).$$

To satisfy the implication rule, 
if $f_{r_1}(h,t) = \mathcal{M}$ then
$f_{r_2}(h,t) = \mathcal{M}$ or if $f_{r_1}(h,t) = 0$ then
$f_{r_2}(t,h) = 0$ or $ f_{r_2}(t,h) = \mathcal{M}$.
We conclude 
$f_{r}(h,t) \leq f_{r}(t,h).$ 

From Equation (1) in the paper, we have:
$$\Phi_{h,t}^T \vec{\beta}^{r_1} \leq \Phi_{h,t}^T \vec{\beta}^{r_2}.$$

We conclude that:
$$\Phi_{h,t}^T (\vec{\beta}^{r_1}  - \vec{\beta}^{r_2}) \leq 0.$$

\end{proof}

\begin{proof}[Proof for the Transitivity Relation.]

To prove transitivity, we can use the truth table of the rule.
Considering Equation (1), we already assume that $f_r(h,t) = \mathcal{M}> 0 $ denotes True and $f_r(h,t) = 0$ denotes False.

In the following conditions, the rule is True:

If $(h,r,t)$ is True, $(t,r,s)$ is True then $(h,r,s)$ is True.

If $(h,r,t)$ is False, $(t,r,s)$ is True, then $(h,r,s)$ is True.

If $(h,r,t)$ is True, $(t,r,s)$ is False, then $(h,r,s)$ is True.

If $(h,r,t)$ is False, $(t,r,s)$ is False, then $(h,r,s)$ is True.

Otherwise, the rule is False. 

The constraint 
$\sigma(\Phi_{h,t} \vec{\beta}^r) \times \sigma(\Phi_{t,s} \vec{\beta}^r)- \sigma(\Phi_{h,s}^T {\vec{\beta}^r}) \leq 0 $
follows the truth table.

\end{proof}

The proof for relations Asymmetric, Negation, Inverse, Reflexive, Irreflexive and Composition are similarly done.





\bibliographystyle{abbrv} 
\bibliography{ijcai19.bbl}

\begin{thebibliography}{}

\bibitem[\protect\citeauthoryear{Akrami \bgroup \em et al.\egroup
  }{2018}]{akrami2018reevaluation}
F.~Akrami, L.~Guo, W.~Hu, and C.~Li.
\newblock Re-evaluating embedding-based knowledge graph completion methods.
\newblock In {\em ACM-CIKM}, 2018.

\bibitem[\protect\citeauthoryear{Bordes \bgroup \em et al.\egroup
  }{2013}]{bordes2013translating}
A.~Bordes, N.~Usunier, A.~Garcia-Duran, J.~Weston, and O.~Yakhnenko.
\newblock Translating embeddings for modeling multi-relational data.
\newblock In {\em NIPS}, 2013.

\bibitem[\protect\citeauthoryear{Demeester \bgroup \em et al.\egroup
  }{2016}]{demeester2016lifted}
T.~Demeester, T.~Rockt{\"a}schel, and S.~Riedel.
\newblock Lifted rule injection for relation embeddings.
\newblock {\em arXiv:1606.08359}, 2016.

\bibitem[\protect\citeauthoryear{Dettmers \bgroup \em et al.\egroup
  }{2018}]{ConvEdettmers2018convolutional}
T.~Dettmers, P.~Minervini, P.~Stenetorp, and S.~Riedel.
\newblock Convolutional 2d knowledge graph embeddings.
\newblock In {\em AAAI}, 2018.

\bibitem[\protect\citeauthoryear{Ding \bgroup \em et al.\egroup
  }{2018}]{SimpConstding2018improving}
B.~Ding, Q.~Wang, B.~Wang, and L.~Guo.
\newblock Improving knowledge graph embedding using simple constraints.
\newblock In {\em ACL}, 2018.

\bibitem[\protect\citeauthoryear{Dong \bgroup \em et al.\egroup
  }{2014}]{ER-MLPdong2014knowledge}
X.~Dong, E.~Gabrilovich, G.~Heitz, W.~Horn, Ni~Lao, K.~Murphy, T.~Strohmann,
  S.~Sun, and W.~Zhang.
\newblock Knowledge vault: A web-scale approach to probabilistic knowledge
  fusion.
\newblock In {\em ACM-SIGKDD}, 2014.

\bibitem[\protect\citeauthoryear{Guan \bgroup \em et al.\egroup
  }{2018}]{SENNguan2018shared}
S.~Guan, X.~Jin, Y.~Wang, and X.~Cheng.
\newblock Shared embedding based neural networks for knowledge graph
  completion.
\newblock In {\em 27th ACM-CIKM}, 2018.

\bibitem[\protect\citeauthoryear{Guo \bgroup \em et al.\egroup
  }{2016}]{guo2016jointly}
S.~Guo, Q.~Wang, L.~Wang, B.~Wang, and Li~Guo.
\newblock Jointly embedding knowledge graphs and logical rules.
\newblock In {\em EMNLP}, 2016.

\bibitem[\protect\citeauthoryear{Guo \bgroup \em et al.\egroup
  }{2018}]{guo2018RUGE}
S.~Guo, Q.~Wang, L.~Wang, B.~Wang, and Li~Guo.
\newblock Knowledge graph embedding with iterative guidance from soft rules.
\newblock In {\em AAAI}, 2018.

\bibitem[\protect\citeauthoryear{Guu \bgroup \em et al.\egroup
  }{2015}]{Relationpathguu2015traversing}
K.~Guu, J.~Miller, and P.~Liang.
\newblock Traversing knowledge graphs in vector space.
\newblock In {\em EMNLP}, 2015.

\bibitem[\protect\citeauthoryear{Han \bgroup \em et al.\egroup
  }{2018}]{TBNNhan2018triple}
X.~Han, C.~Zhang, T.~Sun, Y.~Ji, and Z.~Hu.
\newblock A triple-branch neural network for knowledge graph embedding.
\newblock {\em IEEE Access}, 6, 2018.

\bibitem[\protect\citeauthoryear{Huang \bgroup \em et al.\egroup
  }{2000}]{huang2000classification}
G.B Huang, Y.Q Chen, and H.A Babri.
\newblock Classification ability of single hidden layer feedforward neural
  networks.
\newblock {\em IEEE TNN}, 11(3), 2000.

\bibitem[\protect\citeauthoryear{Kazemi and Poole}{2018}]{SimplE}
S.M Kazemi and D.~Poole.
\newblock Simple embedding for link prediction in knowledge graphs.
\newblock {\em arXiv:1802.04868}, 2018.

\bibitem[\protect\citeauthoryear{Kuttler}{2011}]{kuttler2011multivariable}
Kenneth Kuttler.
\newblock Multivariable calculus, applications and theory.
\newblock 2011.

\bibitem[\protect\citeauthoryear{Lacroix \bgroup \em et al.\egroup
  }{2018}]{lacroix2018canonical}
T.~Lacroix, N.~Usunier, and G.~Obozinski.
\newblock Canonical tensor decomposition for knowledge base completion.
\newblock {\em arXiv:1806.07297}, 2018.

\bibitem[\protect\citeauthoryear{Lin \bgroup \em et al.\egroup
  }{2015a}]{RelationPathlin2015modeling}
Y.~Lin, Z.~Liu, H.~Luan, M.~Sun, S.~Rao, and S.~Liu.
\newblock Modeling relation paths for representation learning of knowledge
  bases.
\newblock In {\em EMNLP}, 2015.

\bibitem[\protect\citeauthoryear{Lin \bgroup \em et al.\egroup
  }{2015b}]{TransR}
Y.~Lin, Z.~Liu, M.~Sun, Y.~Liu, and X.~Zhu.
\newblock Learning entity and relation embeddings for knowledge graph
  completion.
\newblock In {\em AAAI}, volume~15, 2015.

\bibitem[\protect\citeauthoryear{Liu \bgroup \em et al.\egroup
  }{2017}]{liu2017analogical}
H.~Liu, Y.~Wu, and Y.~Yang.
\newblock Analogical inference for multi-relational embeddings.
\newblock In {\em ICML}, 2017.

\bibitem[\protect\citeauthoryear{Minervini \bgroup \em et al.\egroup
  }{}]{minervini2017adversarial}
P.~Minervini, T.~Demeester, T.~Rockt{\"a}schel, and S.~Riedel.
\newblock Adversarial sets for regularising neural link predictors.
\newblock {\em arXiv:1707.07596}.

\bibitem[\protect\citeauthoryear{Minervini \bgroup \em et al.\egroup
  }{2017}]{minervini2017regularizing}
P.~Minervini, L.~Costabello, E.~Mu{\~n}oz, V.~Nov{\'a}{\v{c}}ek, and P.Y
  Vandenbussche.
\newblock Regularizing knowledge graph embeddings via equivalence and inversion
  axioms.
\newblock In {\em ECML PKDD}, 2017.

\bibitem[\protect\citeauthoryear{Neelakantan \bgroup \em et al.\egroup
  }{2015}]{RelationPathneelakantan2015compositional}
A.~Neelakantan, B.~Roth, and A.~McCallum.
\newblock Compositional vector space models for knowledge base completion.
\newblock {\em arXiv:1504.06662}, 2015.

\bibitem[\protect\citeauthoryear{Nguyen \bgroup \em et al.\egroup
  }{2018}]{ConvKBnguyen2017novel}
D.Q Nguyen, T.D Nguyen, D.Q Nguyen, and D.~Phung.
\newblock A novel embedding model for knowledge base completion based on
  convolutional neural network.
\newblock In {\em NAACL-HLT}, 2018.

\bibitem[\protect\citeauthoryear{Nickel \bgroup \em et al.\egroup
  }{2012}]{nickel2012factorizing}
M.~Nickel, V.~Tresp, and H.P. Kriegel.
\newblock Factorizing {YAGO}: scalable machine learning for linked data.
\newblock In {\em 21st conf. on World Wide Web}. ACM, 2012.

\bibitem[\protect\citeauthoryear{Nickel \bgroup \em et al.\egroup
  }{2016}]{Nickel2015ReviewRelationalMLforKG}
M.~Nickel, K.~Murphy, V.~Tresp, and E.~Gabrilovich.
\newblock A review of relational machine learning for knowledge graphs.
\newblock {\em Proc. of IEEE}, 104(1), 2016.

\bibitem[\protect\citeauthoryear{Rockt{\"a}schel \bgroup \em et al.\egroup
  }{2015}]{rocktaschel2015injecting}
T.~Rockt{\"a}schel, S.~Singh, and S.~Riedel.
\newblock Injecting logical background knowledge into embeddings for relation
  extraction.
\newblock In {\em NAACL-HLT}, 2015.

\bibitem[\protect\citeauthoryear{Shi and Weninger}{2017}]{projeshi2017}
B.~Shi and T.~Weninger.
\newblock Proje: Embedding projection for knowledge graph completion.
\newblock In {\em AAAI}, volume~17, 2017.

\bibitem[\protect\citeauthoryear{Socher \bgroup \em et al.\egroup
  }{2013}]{socher2013reasoning}
R.~Socher, D.~Chen, C.D Manning, and Andrew Ng.
\newblock Reasoning with neural tensor networks for knowledge base completion.
\newblock In {\em NIPS}, 2013.

\bibitem[\protect\citeauthoryear{Sun \bgroup \em et al.\egroup }{2019}]{RotatE}
Z.~Sun, Z.~Deng, J.~Nie, and J.~Tang.
\newblock Factorizing yago: scalable machine learning for linked data.
\newblock In {\em ICLR}, 2019.

\bibitem[\protect\citeauthoryear{Trouillon \bgroup \em et al.\egroup
  }{2016}]{ComplEx}
T.~Trouillon, J.~Welbl, S.~Riedel, {\'E}.~Gaussier, and G.~Bouchard.
\newblock Complex embeddings for simple link prediction.
\newblock In {\em ICML}, 2016.

\bibitem[\protect\citeauthoryear{Trouillon \bgroup \em et al.\egroup
  }{2017}]{ComplexJourn}
T.~Trouillon, C.~Dance, {\'E}.~Gaussier, J.~Welbl, S.~Riedel, and et~al.
\newblock Knowledge graph completion via complex tensor factorization.
\newblock {\em JMLR}, 18(1), 2017.

\bibitem[\protect\citeauthoryear{Wang \bgroup \em et al.\egroup
  }{2014}]{TransH}
Z.~Wang, J.~Zhang, J.~Feng, and Z.~Chen.
\newblock Knowledge graph embedding by translating on hyperplanes.
\newblock In {\em AAAI}, volume~14, 2014.

\bibitem[\protect\citeauthoryear{Wang \bgroup \em et al.\egroup
  }{2017}]{wang2017surveyEmbeddings}
Q.~Wang, Z.~Mao, B.~Wang, and Li~Guo.
\newblock Knowledge graph embedding: A survey of approaches and applications.
\newblock {\em IEEE-TKDE}, 29(12), 2017.

\bibitem[\protect\citeauthoryear{Wang \bgroup \em et al.\egroup
  }{2018}]{ProofPaper}
Y.~Wang, R.~Gemulla, and H.~Li.
\newblock On multi-relational link prediction with bilinear models.
\newblock In {\em AAAI}, 2018.

\bibitem[\protect\citeauthoryear{Yang \bgroup \em et al.\egroup
  }{2015}]{Distmult}
B.~Yang, W.t Yih, X.~He, J.~Gao, and Li~Deng.
\newblock Embedding entities and relations for learning and inference in
  knowledge bases.
\newblock In {\em ICLR}, 2015.

\bibitem[\protect\citeauthoryear{Yoon \bgroup \em et al.\egroup
  }{2016}]{LogicTransX}
H.G Yoon, H.J Song, S.B Park, and S.Y Park.
\newblock A translation-based knowledge graph embedding preserving logical
  property of relations.
\newblock In {\em NAACL-HLT}, 2016.

\bibitem[\protect\citeauthoryear{Zhang \bgroup \em et al.\egroup
  }{2019}]{zhang2019quaternion}
Shuai Zhang, Yi~Tay, Lina Yao, and Qi~Liu.
\newblock Quaternion knowledge graph embedding.
\newblock {\em arXiv preprint arXiv:1904.10281}, 2019.

\end{thebibliography}

\end{document}